\documentclass[lettersize,journal]{IEEEtran}
\usepackage{amsmath,amsfonts}
\usepackage{algorithmic}
\usepackage{algorithm}
\usepackage{array}
\usepackage[caption=false,font=normalsize,labelfont=sf,textfont=sf]{subfig}
\usepackage{textcomp}
\usepackage{stfloats}
\usepackage{url}
\usepackage{verbatim}
\usepackage{graphicx}
\usepackage{cite}
\begin{document}


\title{A Hybrid Edge Classifier: Combining TinyML-Optimised CNN with RRAM-CMOS ACAM for Energy-Efficient Inference}

\author{~Kieran~Woodward,~Eiman~Kanjo,~Georgios~Papandroulidakis,~Shady~Agwa,~\IEEEmembership{Member,~IEEE},\\~and~Themis~Prodromakis,~\IEEEmembership{Senior~Member,~IEEE}
\thanks{K. Woodward, is with the Department of Computer Science, Nottingham Trent University, UK \\
E-mail: \{kieran.woodward\}@ntu.ac.uk}%
\thanks{E. Kanjo, is with the Computing Department, Imperial College London, London, UK and the Department of Computer Science, Nottingham Trent University, UK \\
E-mail: \{eiman.kanjo\}@ntu.ac.uk}%
\thanks{G. Papandroulidakis, S. Agwa, and T. Prodromakis are with the Centre for Electronics Frontiers (CEF), Institute for Integrated Micro and Nano Systems, University of Edinburgh, UK \\
E-mails: \{gpapandr, shady.agwa, t.prodromakis\}@ed.ac.uk}%

\thanks{The authors acknowledge the support of the EPSRC FORTE Programme (Grant No. EP/R024642/2), the EPSRC ProSensing Project (Grant No. EP/Y030176/1) and the RAEng Chair in Emerging Technologies (Grant No. CiET1819/2/93).}
}

\markboth{IEEE Transactions on Knowledge and Data Engineering,~Vol., No., February~2025}%
{K. Woodward \MakeLowercase{\textit{et al.}}: A Hybrid Edge Classifier: Combining TinyML-Optimised CNN with RRAM-CMOS ACAM for Energy-Efficient Inference}


\maketitle

\begin{abstract}
In recent years, the development of smart edge computing systems to process information locally is on the rise. Many near-sensor machine learning (ML) approaches have been implemented to introduce accurate and energy efficient template matching operations in resource-constrained edge sensing systems, such as wearables. To introduce novel solutions that can be viable for extreme edge cases, hybrid solutions combining conventional and emerging technologies have started to be proposed. Deep Neural Networks (DNN) optimised for edge application alongside new approaches of computing (both device and architecture -wise) could be a strong candidate in implementing edge ML solutions that aim at competitive accuracy classification while using a fraction of the power of conventional ML solutions. In this work, we are proposing a hybrid software-hardware edge classifier aimed at the extreme edge near-sensor systems. The classifier consists of two parts: (i) an optimised digital tinyML network, working as a front-end feature extractor, and (ii) a back-end RRAM-CMOS analogue content addressable memory (ACAM), working as a final stage template matching system. The combined hybrid system exhibits a competitive trade-off in accuracy versus energy metric with $E_{front-end}$ = $96.23 nJ$ and $E_{back-end}$ = $1.45 nJ$ for each classification operation compared with 78.06$\mu$J for the original teacher model, representing a 792-fold reduction, making it a viable solution for extreme edge applications.
 
\end{abstract}

\begin{IEEEkeywords}
tinyML optimisation, knowledge distillation, analogue memory-centric classifier, RRAM-based content addressable memory
\end{IEEEkeywords}

\section{Introduction}
The proliferation of edge computing devices has created an increasing demand for efficient deployment of Deep Neural Networks (DNN) in resource-constrained environments. Deep neural networks have achieved remarkable success across various applications such as image classification \cite{Rawat2017}, natural language processing \cite{Otter2021} and speech recognition \cite{Mehrish2023} among others. However, their deployment on edge devices remains challenging due to two fundamental bottlenecks, the high computational complexity of fully connected layers and the substantial energy cost associated with memory access operations \cite{Nagaraju2023}. These challenges are particularly notable in the final classification layers of neural networks, where dense matrix operations require numerous memory accesses and floating-point computations. This has led to significant interest in developing methods to compress and optimise neural networks while maintaining acceptable performance levels \cite{Zandonati2023, Li2023a}.

The traditional approach to addressing these challenges has focused on model compression techniques, including pruning, quantisation, and knowledge distillation \cite{Liang2021, Gou2021}. While these methods have shown success in reducing model size and computational requirements, they primarily optimise within the constraints of conventional digital architectures. Such approaches still rely on matrix multiplication operations and maintain the fundamental memory access patterns that contribute significantly to energy consumption. Recent work \cite{Zheng2022a} has demonstrated that memory access operations can consume more energy than arithmetic operations in neural network inference, highlighting the limitations of purely algorithmic optimisation approaches.

Current hardware architectures, particularly in edge computing environments, face limitations in achieving energy efficiency and low-latency performance \cite{Golpayegani2024}. The bottlenecks often arise from the high energy consumption of memory access operations, the computational complexity of fully connected layers, and the reliance on conventional digital processing. In resource-constrained environments, the demand for local data processing necessitates both innovative model optimisations and novel hardware solutions. Emerging analogue hardware accelerators such as RRAM-based architectures, offer a promising avenue for mitigating these challenges by enabling parallel in-memory computation and reducing the energy costs associated with traditional digital architectures.

To address these challenges, we present an architecture and co-design methodology that combines both optimised DNNs alongside emerging analogue hardware accelerators. More specifically, our approach combines advanced DNN model compression techniques with RRAM-based template pattern matching in the analogue domain, replacing the computationally expensive and energy-intensive fully connected layers with more energy efficient comparison operations. Through dynamic knowledge distillation, we transfer the capabilities of a complex neural network to a smaller model specifically designed for template generation. These templates are then used for classification through pattern matching rather than traditional matrix multiplication and activation functions.

This approach offers several key advantages over conventional solutions. First, it eliminates the need for expensive floating-point operations in the dense multi-level perceptron classification stage, replacing them with lower complexity parallel comparison operations. Second, the template-based matching process aligns with emerging hardware architectures optimised for parallel pattern matching, enabling more efficient implementation. Third, our quantisation scheme for template generation is specifically designed to minimise memory requirements while maintaining classification accuracy. Finally, by considering both hardware and software constraints in our design, we achieve a more holistic optimisation that addresses both computational complexity and energy efficiency.

In this work, we are showcasing a mixed-signal approach in designing an energy efficient edge classifier by integrating optimised DNN techniques with ACAM-based pattern matching.

\begin{itemize}
    \item We propose a novel co-design methodology that combines digital DNN optimisation techniques with analogue hardware, enabling energy-efficient and low-latency edge classification.
    \item We introduce a knowledge distillation and pruning framework that compresses a pre-trained DNN into a compact model optimised for edge inference. This approach ensures high accuracy while reducing computational complexity, making it well-suited for deployment on resource-constrained devices.
    \item We develop robust techniques tailored for the generation of binary templates for ACAM-based classification and explore the impact of multi-template strategies and clustering methods on classification accuracy.
\end{itemize}

The remainder of this paper is organised as follows: Section II presents the model optimisation methodology, including knowledge distillation, pruning strategies, and quantisation schemes tailored for deployment on ACAM hardware. Section III introduces the ACAM hardware design, highlighting its memory-centric architecture and energy-efficient template matching capabilities, Section IV describes the experimental setup, Section V evaluates the results and finally Section VI concludes.

\section{Model Optimisation}
In this section, we present our methodology for developing a compressed neural network model optimised for deployment on Analogue Content-Addressable Memory (ACAM) hardware. Our approach combines state-of-the-art model compression techniques with optimisations to create an efficient and edge computing hardware-friendly model. 

Our proposed approach consists of several stages designed to contribute to the overall goal of creating an ACAM-optimised model:

\begin{enumerate}
\item Knowledge Distillation: A teacher-student framework is utilised, where a larger, more complex model (the teacher) is used to train a smaller, more compact model (the student). This technique enables the development of a model that maintains high accuracy while significantly reducing computational requirements and memory footprint.
\item Pruning: Following distillation, an iterative pruning strategy is applied to further reduce the model size. This step involves identifying and removing less important weights and connections, resulting in a sparser network structure.
\item quantisation: We implement a quantisation scheme that reduces the precision of weights and activations. 
\item ACAM optimisations: Utilising the compressed model, we generate templates suitable for pattern matching, designed to utilise the analogue computation capabilities of ACAM hardware.

\end{enumerate}

\subsection{Knowledge distillation}
Knowledge distillation forms a crucial component of the methodology for creating smaller, efficient models \cite{Hinton2015}. Our implementation follows a teacher-student framework where knowledge is transferred from a high-capacity teacher network to a more compact student network.

The knowledge distillation process is controlled by two key parameters: $\alpha$ and temperature $T$. The $\alpha$ parameter balances the importance between learning from the teacher's soft predictions and learning from the ground truth labels. The temperature parameter $T$ controls the softness of probability distributions during knowledge transfer, where higher temperatures produce softer distributions that reveal more fine-grained knowledge about the relationships between the classes that the teacher has learned.

Traditional supervised learning typically uses \textit{hard} labels for training. However, these hard labels don't capture the rich information about similarities between classes that the teacher model has learned. Knowledge distillation preserves this information by using the teacher's \textit{soft} probability distributions as training input for the student model.

The knowledge distillation process is guided by a composite loss function that combines two components:

\begin{equation} \label{kd}
L = \alpha L_{KD}(z_s, z_t) + (1-\alpha)L_{CE}(z_s, y)
\end{equation}

\noindent
where $L_{KD}$ represents the knowledge distillation loss between the student outputs $z_s$ and teacher outputs $z_t$, $L_{CE}$ denotes the standard cross-entropy loss between student predictions and ground truth labels $y$, and $\alpha$ is the balancing parameter.

The knowledge distillation loss $L_{KD}$ is computed using the Kullback-Leibler (KL) divergence between the softened probability distributions of the student and teacher models:

\begin{equation} \label{kd_loss}
L_{KD}(z_s, z_t) = T^2KL(\sigma(z_s/T) \parallel \sigma(z_t/T))
\end{equation}

\noindent
where $T$ is the temperature parameter that controls the softness of the probability distributions, and $\sigma$ typically represents the activation function:

\begin{equation} \label{kd_temp}
\sigma(z_i) = \frac{\exp(z_i/T)}{\sum_j \exp(z_j/T)}
\end{equation}

The temperature scaling in equation \ref{kd_temp} produces softer probability distributions by reducing the magnitudes of the logits, revealing more information about the inter-class relationships learned by the teacher. The $T^2$ scaling factor in equation \ref{kd_loss} ensures the gradient magnitudes remain approximately constant when changing the temperature, as initially proposed by Hinton et al \cite{Hinton2015}.

To improve the effectiveness of knowledge transfer, we incorporate curriculum learning by sorting the training data based on the teacher model's confidence. For each training sample $(x_i, y_i)$, we calculate a difficulty score $d$:

\begin{equation}
d(x_i, y_i) = L(z_t(x_i), y_i)
\end{equation}

\noindent
where $z_t(x_i)$ is the teacher's prediction for input $x_i$, and $L$ represents the cross-entropy loss between the prediction and true label $y_i$. The training data is then ordered from easiest (lowest loss) to hardest (highest loss), allowing the student to gradually progress from simple to more challenging examples during training.

This knowledge distillation framework enables our student model to benefit from the rich representations learned by the larger teacher network while maintaining a significantly smaller parameter count suitable for edge deployment. The combination of soft targets and curriculum learning helps the student model learn more effectively, leading to better generalisation despite its reduced capacity.

\subsection{Pruning strategy}
After knowledge distillation, we apply structured pruning to further reduce the model size and computational requirements while maintaining performance. We employ a magnitude-based pruning approach that systematically removes less important weights based on their absolute values, followed by fine-tuning to improve accuracy \cite{Yang2020}.

The pruning schedule follows a polynomial decay that gradually increases network sparsity from an initial value $s_i$ = 0.50 (50\% sparsity) to a final target sparsity $s_f$ = 0.80 (80\% sparsity). This aggressive pruning strategy was chosen to substantially reduce model complexity while maintaining acceptable performance through the iterative pruning and fine-tuning process. The sparsity at step $t$ is calculated as:

\begin{equation}
s(t) = s_f + (s_i - s_f)(1 - \frac{t}{n_t})^3
\end{equation}

At each pruning step, weights are ranked according to their absolute magnitude:

\begin{equation}
r(w_{ij}) = |w_{ij}|
\end{equation}

\noindent
where $w_{ij}$ represents the weight connecting neurons $i$ and $j$. Weights with the lowest magnitude are considered less important and are candidates for pruning. The pruning threshold $\theta_t$ at step $t$ is determined by:

\begin{equation}
\theta_t = Q(|W|, s(t))
\end{equation}

\noindent
where $Q(|W|, p)$ returns the $p$-th percentile of the absolute weights $|W|$, and $s(t)$ is the target sparsity at step $t$. Any weight with magnitude below this threshold is set to zero.


To maintain model performance, we implement an iterative pruning and fine-tuning schedule. After each pruning step, the network undergoes a brief period of fine-tuning to allow the remaining weights to adapt and compensate for the pruned connections. This process continues until the final target sparsity is reached.

After the pruning process is complete, a final fine-tuning phase is performed to ensure the pruned model maintains its accuracy. The remaining non-zero weights are then stored using a sparse matrix format, significantly reducing the model's memory footprint. The combination of gradual sparsity increase and iterative fine-tuning helps preserve the most important features learned by the network while eliminating redundant parameters. This compressed format is particularly well-suited for deployment on the edge, as it reduces both storage requirements and computational complexity during inference.

\subsection{Quantisation scheme}
Our quantisation strategy operates in two stages: first, quantizing the model weights using quantisation-aware training, and second, quantizing the feature maps to align with hardware constraints.

The model weights are quantised to 8-bit integers during training. This process applies quantisation during training, allowing the model to adapt to reduced precision, while still helping to preserve accuracy compared with lower bit quantisation during training. For ACAM deployment, we further quantise the feature maps. We use a mean-based thresholding approach for binary quantisation. This approach, using the mean rather than a fixed threshold, better adapts to the distribution of feature values.









\subsection{ACAM-aware optimisations}
Several architectural decisions in our network design are specifically guided by ACAM hardware constraints and efficiency considerations. The primary optimisation focuses on minimising the use of complex layer types and reducing the overall network depth while still aiming to maintain performance.

A key design decision is the exclusion of fully connected layers throughout the network. Traditional neural networks often employ multiple fully connected layers, particularly in their final stages, which significantly increases the parameter count and computational complexity. Instead, our architecture relies primarily on convolutional layers, which are more parameter-efficient due to weight sharing and maintain spatial relationships in the feature maps while also demonstrating superior memory access patterns. In convolutional operations, each loaded weight is reused multiple times as it slides across the input feature map and input values are reused across multiple filter applications. This repeated use of data once loaded from memory significantly improves the energy efficiency of the network, as the energy cost of memory access is spread over multiple computations. This design choice not only reduces the model size but also aligns better with ACAM's pattern matching capabilities. 

The number of convolutional layers is also carefully optimised. Rather than utilising deep networks with numerous convolutional layers, we employ a compact design with strategically placed convolutional layers. By using a progressive expansion and reduction in filter count we aim to maintain essential feature extraction capabilities while minimising the overall computational requirements. The resultant lightweight model is then used for the template generation and pattern matching that replaces the classification functionality of a traditional softmax layer.

\subsubsection{Template generation}
Template generation is a crucial step in optimising neural networks for deployment on ACAM hardware \cite{Aguirre2024}. This process transforms the learned representations of a trained neural network into a format that can be efficiently stored and matched within the ACAM architecture. The primary goal of template generation is to create a set of representative patterns for each class that capture the features learned by the network.

Template generation aims to distill the high-dimensional feature representations learned by the neural network into a more compact and hardware-friendly format. This is achieved by analysing the logits of the network's penultimate layer (before SoftMax classification) when processing training samples from each class. These logits are then processed to create templates that capture the most important features for distinguishing between classes.

The trained neural network is used to generate feature maps for a large number of training samples from each class. These feature maps represent the network's internal representation of the input data. 

Template generation is a crucial step in optimising neural network models for ACAM hardware implementation. This process involves converting the feature maps produced by the neural network into binary representations that can be efficiently stored and processed in ACAM architecture. Two primary methods for this binary thresholding have been explored; mean-based and median-based approaches. 

\begin{figure}[h]
    \centering
    \includegraphics[width=0.95\linewidth]{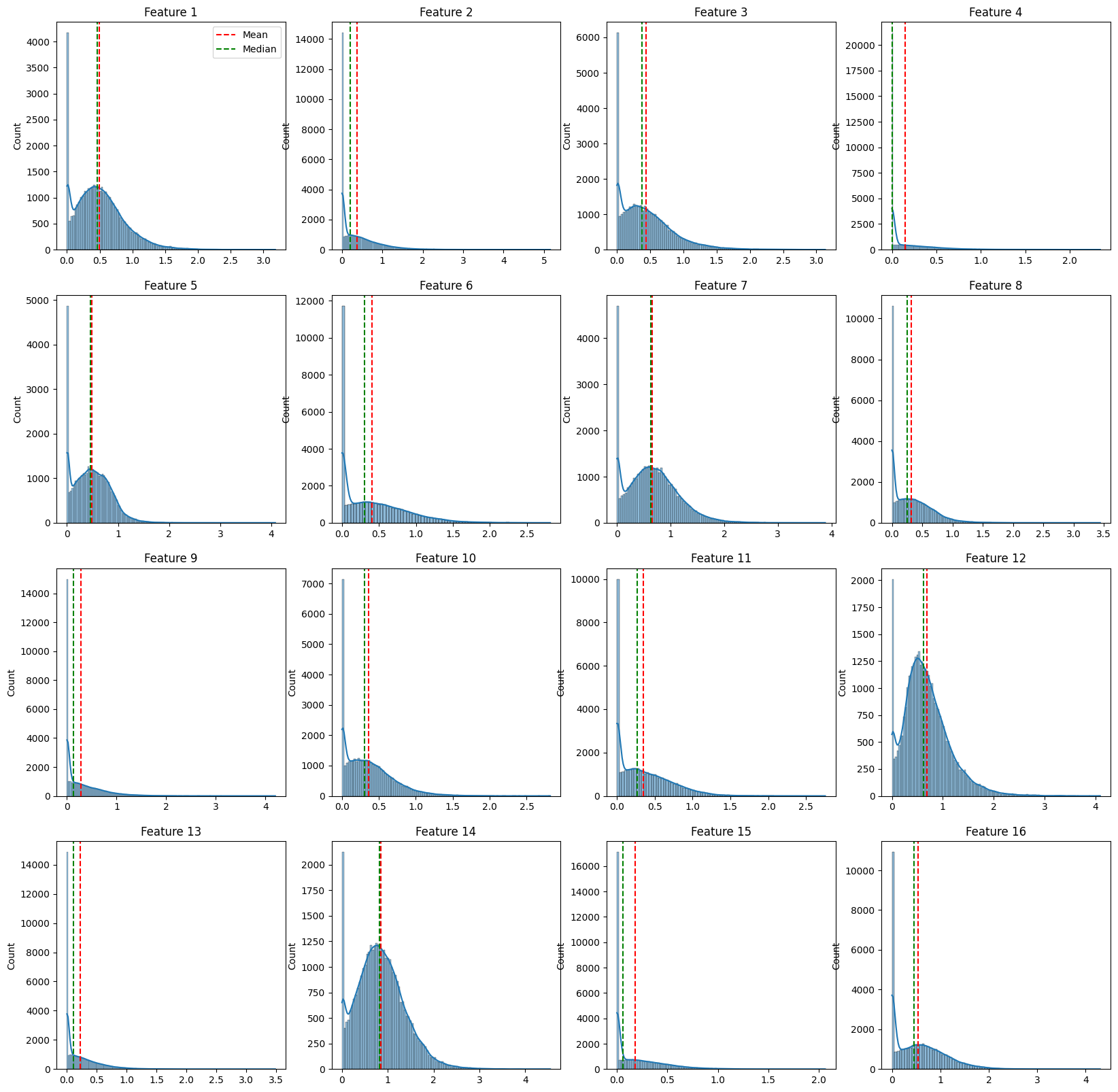}
    \caption{Comparison of mean (red) and median (green) thresholding for each feature output from the front-end classifier.}
    \label{fig:feature_threshold}
\end{figure}

In conventional signal processing, median-based thresholding is often preferred for its robustness to outliers and ability to maintain a balanced distribution of values in the binary templates \cite{Kantharimuthu2024, Burman2024, Wang2011a}. However, our analysis (see Fig. \ref{fig:feature_threshold}) reveals that for neural network feature maps, mean-based thresholding consistently outperformed the median approach. In the mean-based method, the average value of each feature across all training samples is calculated to serve as the threshold, with feature values above the mean set to 1 and those below set to 0. While this approach is traditionally considered more sensitive to outliers, it can capture more nuanced distributions in the feature space which is advantageous in the context of neural network activations. Neural networks, particularly those employing ReLU activations, produce feature maps characterised by significant sparsity, with numerous zero values. These zero activations substantially lower the mean threshold compared to the median, creating a more discriminative binary representation. The lower threshold enables the binary templates to capture subtle, non-zero activations that often carry important class-specific information, whereas a median-based approach might eliminate these features by setting a threshold that effectively treats many informative, low-magnitude activations as noise. Therefore, the mean-based method is more sensitive to the actual distribution of active features, preserving more relevant information for classification.

We also explore the possibility of generating multiple templates for each class, specifically exploring configurations with one, two, or three templates per class. This approach acknowledges the variability within class distributions and aims to capture a more comprehensive representation of each class's feature space. By creating multiple templates, we can encompass different sub-clusters within a class, potentially leading to improved classification accuracy. The multi-template strategy is particularly beneficial for classes with high intra-class variability. In our implementation, these multiple templates are generated using a clustering approach on the feature representations of each class. For instance, when using three templates per class, we apply k-means clustering with k=3 on the class-specific feature maps, and the centroids of these clusters serve as the templates. K-means was chosen for its computational efficiency and ability to partition data into spherical clusters, making it a well-suited for this task. During the matching process, an input query is compared against all templates for each class, and the best match among these templates determines the class similarity score. 

To further optimise the clustering process, we utilised silhouette scores to evaluate the quality of clustering for different configurations. Silhouette scores provided a quantitative measure of how well-separated and cohesive the clusters are, enabling the selection of the number of templates that best captured the intra-class variability. This ensures that the generated templates represent meaningful and distinct sub-clusters within each class. While the multi-template approach increases the memory footprint and the number of comparisons required, it offers a trade-off between model complexity and representational power, potentially leading to more robust and accurate classifications, especially for complex datasets with significant intra-class variations. 

\subsubsection{ACAM Pattern Matching Computing Concept} \label{sec:pattern_matching_concept}
We propose using pattern matching to serve as the inference mechanism, replacing the traditional final classifcation function of conventional digital DNNs with a hardware-optimised similarity search operation. The main principle of pattern matching is to compare an input pattern against a set of stored templates. In this work this is performed by a RRAM-CMOS ACAM system which is discussed in Section \ref{sec:ACAM_hardware}. To evaluate the performance of the hybrid software-hardware classifier, the modelling of the ACAM is required towards introducing a good approximation of its behaviour to the software flow. Additionally, due to the intrinsic properties of the RRAM devices, the employment of program-once-read-many can provide better stability and endurance of the emerging RRAM devices. Thus, the calibration of the appropriate weights that are required for the RRAM-based ACAM in software and the programming once to the hardware is considered a more pragmatic method.
For inference (assuming that the appropriate weights have been programmed to ACAM during training phase), the input pattern typically represents the feature map generated by processing an input sample through the front-end convolutional feature extractor network. This feature map is used as a query key to be compared against all stored templates. The stored templates have been generated during the training process to be used as the learned representation classes to be classified by the network. The comparison, which can be performed in parallel for all stored templates, identifies the template that most closely matches the input pattern, therefore determining the classification output by selecting the template that resulted in the highest similarity.
A simple feature count pattern matching model is first explored to emulate the main functionality of ACAM without introducing the more hardware-demanding distance calculation from the matching window. Thus, unlike a more complex similarity calculation model, we are assuming that if an input is outside the effective matching window of each ACAM cell then it's effect to the output is zero. Each input sample's quantised feature map is compared against the stored templates for all classes. The comparison sums the number of exactly matching features between the input and each template. This method is chosen for its simplicity and efficiency in hardware implementation, as it primarily involves bitwise comparison operations that can be performed efficiently. The class with the highest feature count is selected as the predicted class for the input sample. 

For feature count-based pattern matching, given a query feature vector Q and template T, each with N features, the feature count score $S_{fc}$ can be expressed as:

\begin{equation}
S_{fc}(Q,T) = \sum_{i=1}^{N} I(Q_i = T_i)
\end{equation}

\noindent
where $I$ is the indicator function and $\epsilon$ is the matching threshold.
For a more realistic model of ACAM, we are also testing a similarity-based pattern matching approach which implements a distance calculation when the input is outside the matching window of the ACAM cell. For each template, the function calculates a similarity score based on how well the query's features fall within the template's defined bounds. Specifically, it computes both a distance measure and a hit ratio. The distance is calculated as the sum of squared differences between the query and the template bounds, but only for features that fall outside the attribute's matching range. This approach allows for a degree of flexibility in pattern matching, accommodating small variations in feature values. The hit ratio represents the proportion of features that fall within the template's bounds, providing a measure of overall match quality. The final similarity score is derived from these metrics, with higher similarities indicating closer matches. This similarity-based approach provides a more granular assessment of match quality compared to the feature count method, potentially offering improved discrimination between closely related patterns. 

For similarity-based pattern matching, we calculate both distance and hit ratio. Given a template T with upper and lower bounds $[T^L, T^U]$, the distance score $D$ for features outside the template bounds is:

\begin{equation}
D(Q,T) = \sum_{i=1}^{N} \begin{cases} 
(Q_i - T^U_i)^2 & \text{if } Q_i > T^U_i \\
(T^L_i - Q_i)^2 & \text{if } Q_i < T^L_i \\
0 & \text{otherwise}
\end{cases}
\end{equation}

The hit ratio H measures the proportion of features falling within the template bounds:

\begin{equation}
H(Q,T) = \frac{1}{N}\sum_{i=1}^{N} \mathbb{1}(T^L_i \leq Q_i \leq T^U_i)
\end{equation}

The final similarity score combines both metrics:

\begin{equation}
S_{sim}(Q,T) = \frac{H(Q,T)}{1 + \alpha D(Q,T)}
\end{equation}

\noindent
where $\alpha$ is a scaling parameter that controls the influence of the distance penalty.

For both methods, the final classification decision C for a query Q across M classes is determined by:

\begin{equation}
C(Q) = \underset{j \in \{1,...,M\}}{\arg\max} S(Q,T_j)
\end{equation}

\noindent
where $S$ represents either $S_{fc}$ or $S_{sim}$ depending on the chosen matching method.

\section{RRAM-CMOS ACAM Hardware Design and Operation} \label{sec:ACAM_hardware}
In this section, we are introducing the hardware description of the RRAM-CMOS ACAM system used to replace the final dense layers in conventional DNNs. The specific technology employed for this ACAM is called Template piXeL (TXL) and depending on the specific requirements of each edge application, different circuit and/or system implementations can be used. Each and every implementation discussed is used to implement specifically the last similarity comparison back-end system, as shown in Fig. \ref{fig:ANN_ACAM_Block}.

\subsection{ACAM Architecture and Background}  \label{sec:ACAM_Background}
Template matching hardware accelerators can be used as energy efficient classification engines closely integrated with edge sensors. Recently, content addressable memory (CAM) is a popular system architecture that has been repeatedly deployed as template matching accelerators \cite{Graves2020}. CAMs are memory-centric systems using custom dense memory arrays with hardware-level optimisation aimed at performing a fast parallel search and match operation between a query key and the stored templates. CAMs operate in a massively parallel manner, comparing the input with every stored templates in the CAM array simultaneously \cite{Graves2020}. In recent years, towards further reducing the energy dissipation of CAM systems, emerging memory technologies have been proposed as strong candidates to implement analogue memory and computing technologies. Some of these advances resulted in Analogue CAMs (ACAMs) that can be used to directly interface with analogue sensors' output and process information on-the-fly without analogue-to-digital conversion that is costly in terms of energy and IC area \cite{Li2020_HP_ACAM}. Implementing analogue template matching accelerator near the edge sensors can further help with data transfer bottlenecks since the information can be efficiently processed using real-valued data locally to the sensor and only the classification results will need to be transmitted. \par
\begin{figure*}[t]
    \centering
    \includegraphics[width=\linewidth]{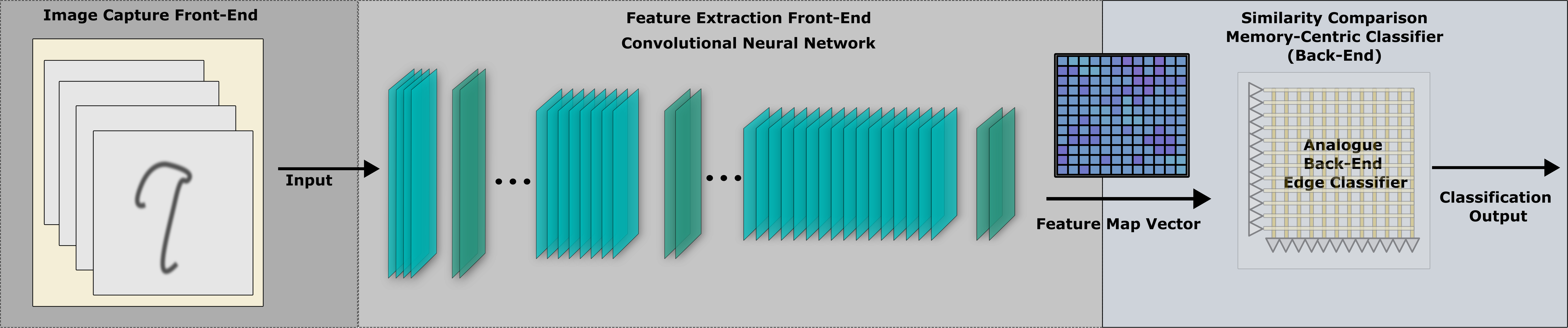}
    \caption{Combined Deep Neural Network (DNN), for feature extraction and dimensionality reduction, with Analogue Content Addressable Memory (ACAM) back-end network \cite{papandroulidakis2024_9t4r_arXiv}, for final classification of the extracted feature map through analogue information processing.}
    \label{fig:ANN_ACAM_Block}
\end{figure*}
On the device engineering front, recent advances in emerging memory technologies introduce a new component for novel analogue circuit design. Memristors, also known as Resistive Random Access Memory (RRAM), are two-terminal, tuneable, non-volatile, nanoscale resistive memory devices \cite{Ielmini2018}. RRAM has many beneficial traits over conventional memory devices. RRAM can be integrated to implement novel computing systems showcasing low power consumption, high throughput, low area of integration and multi-bit information stored per cell compared to their fully CMOS counterparts. \par
In the last decade, RRAM has been extensively investigated for its use in processing information in the analogue domain while being organised in dense topologies. RRAM devices can be used to store multiple bits of information while retaining the stored data in the absence of power, thus enabling dense non-volatile memory modules. The investigation of RRAM devices as a form of low power tuneable resistive element drives analogies with the observed behaviour of biological synaptic connections. Hence, novel DNNs using RRAM-CMOS hybrid circuits and systems have been proposed to satisfy the computing needs, especially for resource-constrained edge applications \cite{Sebastian2020, Payvand2018}. Furthermore, the integration of RRAM devices into CAM architectures can result in novel energy-efficient ACAMs \cite{Pedretti2021}. RRAM-CMOS ACAMs have been shown to be leveraged as an efficient component for ML models \cite{Pedretti2022}. \par
Towards integrating such analogue template matching classifiers in a wide variety of applications that require the classification of high-dimensional data, the use of a feature extractor front-end is required. A concept schematic of such a system is shown in Fig. \ref{fig:ANN_ACAM_Block}. The feature extractor performs the necessary dimensionality reduction by generating a compressed feature map of the initial high-dimensional input. As an example, we could consider the use of an image sensor capable of capturing a still image with a specific number of sensing pixels with a convolutional network being used to perform a spatial feature extraction of the input. The flattened feature map (e.g. the output of a convolutional-based feature extractor) is used as a query input for the back-end RRAM-CMOS ACAM classifier. The classifier is used to perform the final classification of the lower-dimensional feature map by comparing it to its pre-stored feature map templates. The stored templates are created during the network training and are stored in ACAM for inference. The back-end classifier responds with a real-valued output vector that indicates the similarity of the feature map with each of the stored template/landmark feature maps, with the highest similarity being selected for the classified pattern. The similarity search is performed using real-valued data which circuit-wise is implemented by the analogue ACAM layer. Through this configuration of mixed-signal components, we can exploit the benefits of both digital and analogue information processing for ML applications. It has been shown that efficient digital feature extraction system can be implemented through tinyML techniques that can achieve competitive accuracy while using only a small fraction of resources compared to their full DNN counterparts \cite{Gibbs2023a}. Towards further optimising such techniques for extreme edge cases an analogue RRAM-based classifier can be used as final classification stage to replace the dense Multi-Layer Perceptron (MLP) network that conventionally is used for the final fully connected layers in many DNN models. The back-end classifier can process data using the real-valued format of the feature map and can perform the similarity check in a massively parallel manner, thus saving both memory-computing resources as well as inference latency. Depending on the RRAM technology and the ACAM design employed, different weight quantisation formats can be used. The use of RRAM devices enables the integration of multi-bit memory technology near the computing, thus further saving memory resources and inference latency.\par

\subsection{Memory-Centric Accelerators for Energy Efficient Template Matching} \label{sec:TXL_ACAM}
In this work, we are employing specifically Template piXeL (TXL) ACAM (TXL-ACAM) technology, which implements an energy-efficient ACAM system for accelerating near-sensor template matching operations. The ACAM system is designed in a commercially available 180 nm CMOS technology while the RRAM devices are added through back-end-of-line (BEOL) integration. The RRAM devices are designed and fabricated through in-house facilities \cite{Stathopoulos2018}. The RRAM devices per cell are used to define the matching window, thus the voltage range for which an input is considered as matching. The TXL-ACAM implementation showcase competitive energy efficiency compared to other state-of-art ACAM technologies of approximately $185 fJ$ per similarity search operation per cell. \par
\begin{figure}
    \centering
    \includegraphics[width=\linewidth]{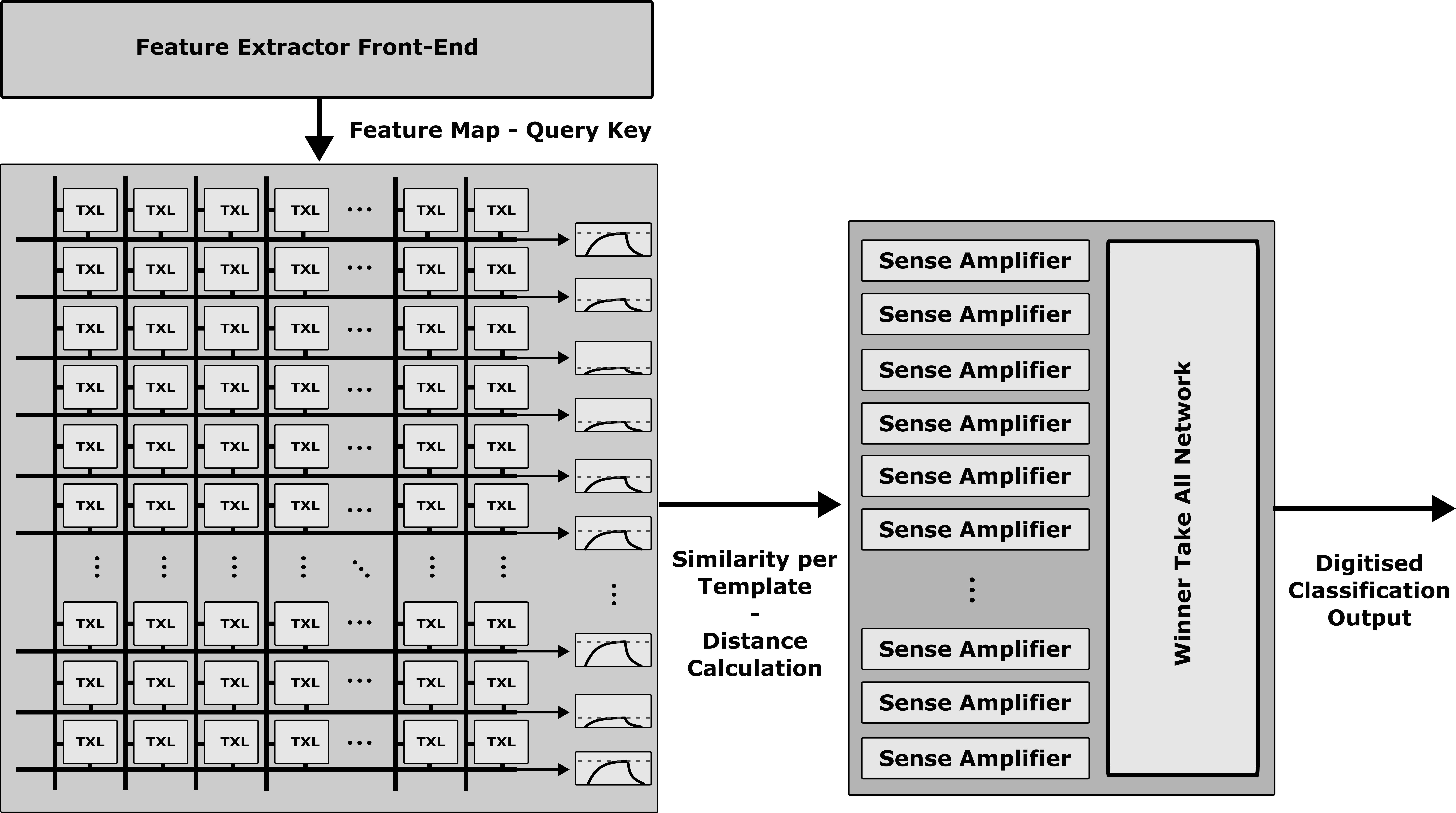}
    \caption{Analogue Content Addressable Memory (ACAM) block diagram showcasing the main analogue computing blocks \cite{papandroulidakis2024_9t4r_arXiv}. The block is effectively a two layer network, with the first layer being the ACAM module that calculates in parallel the similarity of the input feature maps with the pre-computed templates. The second layer senses the similarities and converts them into the proper voltage levels to pass through a final Winner Take All (WTA) network that computes the argmax function on the set of similarities. We assume that the feature map used as input to ACAM is the output of the front-end feature extractor with the ACAM being employed as a final layer classification network. }
    \label{fig:ACAM_Block}
\end{figure}
The TXL-ACAM can classify a low dimensionality input (e.g. the feature map at the output of a convolutional feature extractor) by calculating the distance of the input vector to stored templates. A block diagram of the proposed TXL-ACAM is shown in Fig. \ref{fig:ACAM_Block}. The output of the TXL-ACAM is a real-valued/analogue vector that encodes the similarity of the input per template. This computationally cornerstone operation could be considered similar to the search and match of a query inside a miniaturised database. The set of similarities is processed through a Winner Take All (WTA) network to calculate efficiently the maximum similarity in the analogue domain. The WTA implements the argmax function and provides a one-hot encoding of the highest similarity template. All operations, with regard to the TXL-ACAM system, are implemented in the analogue domain using custom analogue circuitry. This results in an analogue back-end classifier that can replace a larger last stage CMOS-based MLP network used conventionally for the last few stages of DNNs. \par
\begin{figure}
    \centering
    \includegraphics[width=\linewidth]{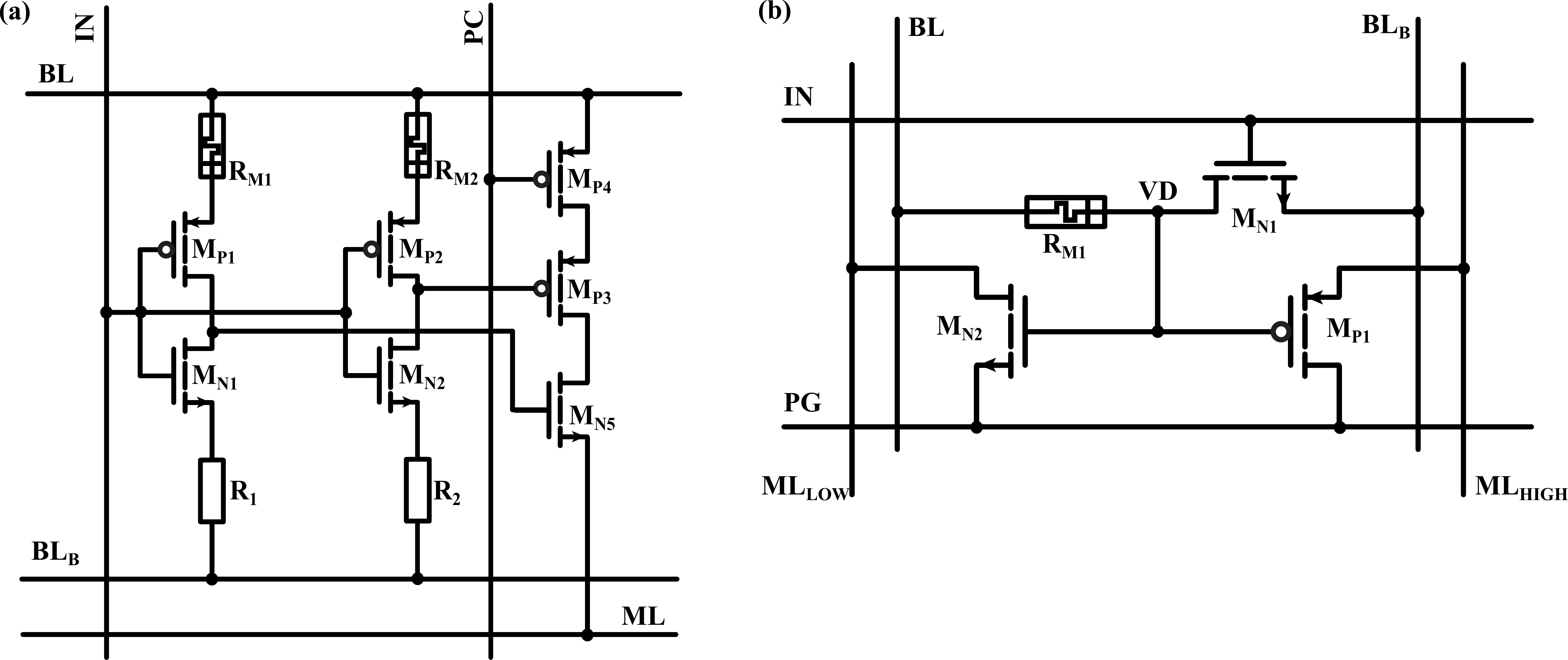}
    \caption{RRAM-CMOS -based Template piXeL (TXL) ACAM cell schematics. There are many version of the TXL-ACAM technologies with each version comprised of a specific set of trade-offs. In (a) a 6T4R charging design is showcased aimed at ML applications with increased sparsity \cite{papandroulidakis2024_9t4r_arXiv} In (b), a 3T1R precharging design is shown that is aimed at applications that has strict area specifications as well as differentiability as trait for the final stage classification \cite{Agwa2023_3T1R_ACAM}.}
    \label{fig:TXL_Circuits}
\end{figure}

 
In Fig. \ref{fig:TXL_Circuits}(a), a schematic representation of the 6T4R TXL-ACAM pixel is shown \cite{papandroulidakis2024_9t4r_arXiv}. The upper and lower thresholds of the matching window per cell are effectively defined based on the ratio of the upper and lower RRAM device which shifts the voltage threshold of the hybrid RRAM-CMOS inverters. Additional circuitry implemented per cell is used to conditionally charge the matchline when the input falls within the cell's matching window. An additional pMOS devices is controlling the cell's output through current limiting. The current limiter pMOS devices is used to calibrate the cell's charging rate in case of a match. The circuit performs a pattern matching operation between its input signal and its stored patterns (stored in the form of programmable conductances through the integration of the non-volatile RRAM devices). Each hybrid inverter stores one of the bounds through appropriately configured RRAM devices. The charge for each cell is accumulated using an analogue capacitor-based integrator circuit implemented for each row of the memory topology (thus per template). If multiple cells have a match with attributes of the query input then multiple connections to $V_{DD}$ are enabled and the matchline is driven at some specific rate (dependent on the number of match enable available per TXL-CAM array) to high voltage. The sense amplifiers can be used to detect if the overall match activity per template exceeds a specific predefined level controlled by the amplifiers threshold voltage. The sense amplifiers are calibrated to detect a specific voltage level which is translated to match based on the time-to-charge dynamics of the matchline. In case no such voltage level is observed within the readout operation timeframe, then the template is considered to not match with the input query at a sufficient rate. The sense amplifiers voltage threshold can be arbitrarily set depending on the intrinsic RRAM-CMOS cell dynamics that define the charging rate of the matchline. \par
In Fig. \ref{fig:TXL_Circuits}(b), a schematic representation of the 3T1R TXL-ACAM pixel is shown \cite{Agwa2023_3T1R_ACAM}. The 3T1R ACAM cell consists of a single 1-transistor-1-resistor (1T1R) memory cell (serial connection of $R_{M1}$ RRAM and $M_{N1}$ nMOS devices) and two additional transistors ($M_{P1}$ pMOS and $M_{N2}$ nMOS) for matchlines evaluation. The input voltage controls the resistance of the $M_{N1}$ transistor, and thus the voltage divider formed by the RRAM devices and the $M_{N1}$. The additional nMOS and pMOS devices ($M_{N1}$ and $M_{P1}$, respectively) are connected to a dual matchline configuration ($ML_{LOW}$ and $ML_{HIGH}$). The two matchlines $ML_{HIGH}$ and $ML_{LOW}$ represent the high and low bounds of the matching window, respectively. Due to the complementary behaviour between nMOS and pMOS, the nMOS transistor is responsible for discharging $ML_{LOW}$ when the input voltage is below the lower bound while the pMOS transistor is responsible for discharging $ML_{HIGH}$ when the input voltage exceeds the higher bound. Contrary to Fig. \ref{fig:TXL_Circuits}(a), the 3T1R design follows a precharging and evaluating operation cycle instead of a discharging/initialisation and evaluation. The larger charging cell could be preferable in sparse activation applications, while the more conventional 3T1R precharging cell design could be preferred in normally distributed activation applications due to its smaller size. \par
More specifically, if the input voltage $IN$ is below the lower bound $V_{LOW}$, the resistance of the nMOS transistor $M_{A}$ is relatively high which drives $VD$ (voltage at the intermediate node of the 1T1R voltage divider circuit) to an appropriate voltage to enable discharge through the nMOS transistor $M_{MLO}$ of matchline $ML_{LOW}$. At the same time, the pMOS transistor $M_{MHI}$ is non-conductive and the matchline $ML_{HIGH}$ is not discharging through the pMOS device.
If the input voltage $IN$ exceeds the higher threshold, the resistance of the nMOS transistor $M_{A}$ is relatively low which drives $VD$ to an appropriate voltage to make the pMOS transistor $M_{MHI}$ $ON$ to discharge the matchline $ML_{HIGH}$ while the nMOS transistor $M_{MLO}$ is $OFF$ and the matchline $ML_{LOW}$ is not discharging fast enough.
If the input voltage $IN$ is lower than the high threshold and higher than the low threshold, then a match case occurs. The safe $VD$ value makes both pMOS transistor $M_{MHI}$ and nMOS transistor $M_{MLO}$ partially or completely $OFF$ so that the matchlines $ML_{HIGH}$ and $ML_{LOW}$ are not discharging. By evaluating both matchlines a final decision about the match/mismatch case can be asserted. Furthermore, being able to evaluate each threshold separately provides better assessment in case of mismatching and makes the cell differentiable. Thus, we can understand how to train the RRAM weights by assessing which threshold is surpasses or not. \par

\section{Experimental Setup}
\subsection{Datasets preparation}
The CIFAR-10 dataset \cite{krizhevsky2009learning} has been used to evaluate model performance as it is a widely recognised benchmark in the field of image classification. This dataset consists of 60,000 32x32 colour images across 10 classes, with 6,000 images per class. The dataset is divided into 50,000 training images and 10,000 test images.

CIFAR-10 serves as a common challenge in image classification tasks, allowing for direct comparison with other state-of-the-art methods. The dataset's size allows for rapid experimentation and iteration, which is crucial when developing new classification approaches. Furthermore, the 10 distinct classes (airplane, automobile, bird, cat, deer, dog, frog, horse, ship, truck) provide a good representation of real-world object recognition tasks.

To better simulate the input characteristics of ACAM hardware we pre-processed the CIFAR-10 images by converting them to grayscale. This conversion was performed using the standard formula: Y = 0.2989 $\times$ R + 0.5870 $\times$ G + 0.1140 $\times$ B, where Y is the resulting grayscale value, and R, G, and B are the red, green, and blue channel values respectively. The grayscale conversion decreases the input dimensionality by a factor of three aligning well with the typical input constraints of ACAM hardware. After the grayscale conversion, the values are normalised to improve the stability of neural network training and the subsequent quantisation process.

\subsection{tinyML Feature Extractor Model architectures}
A ResNet-50 \cite{He2016} architecture is used as the teacher model. This deep residual network is chosen for its high performance on the CIFAR-10 dataset and its ability to learn rich, hierarchical features \cite{Kaushik2024}. The ResNet-50 model consists of three stages, each containing multiple residual blocks. The first stage operates on 16 channels, with subsequent stages doubling the number of channels while reducing spatial dimensions. Each residual block comprises of two 3$\times$3 convolutional layers with batch normalisation and ReLU activation function. A shortcut connection is added to each block, with 1$\times$1 convolutions used when necessary to match dimensions. The network concludes with global average pooling and a fully connected layer for classification. The ResNet-50 model is initialised with He normal initialisation and employs L2 regularisation to prevent overfitting.

\subsubsection{Student Ablation Studies}
Our ablation studies quantify the impact of each optimisation technique and architectural decision on classification performance of the student model. The study explored a range of architectures, from simple dense networks (390,000 parameters) to complex multi-layer convolutional models (5.9M parameters), exploring various configurations including dense layers (128-1024 units), convolutional layers (16-512 filters), and global average pooling. The sequential application of optimisation techniques demonstrated clear benefits. Knowledge distillation consistently improved accuracy across all model configurations, with an average improvement of 5.2\% compared to baseline training. Notably, for the CNN-based architectures, knowledge distillation yielded even more substantial gains, with improvements ranging from 7.3\% to 9.4\%. The subsequent application of pruning maintained this enhanced performance, showing minimal degradation (average -0.8\%) while significantly reducing model complexity. 8-bit training aware quantisation also proved extremely robust, reducing accuracy by only 0.42\% on average while enabling hardware-efficient implementation.

Starting with a baseline model containing dense layers (Dense 128), we observed that increasing dense layer width (256, 512) actually degraded performance, with accuracies dropping to 78.2\% and 77.89\% respectively for binary similarity-based matching. This suggests that larger dense layers may introduce redundant parameters without capturing additional discriminative features. The final architecture, avoiding dense layers to align with ACAM hardware constraints, achieved the optimal balance between model complexity and accuracy with convolutional termination layers outperformed dense layer configurations in terms of stability across quantisation schemes. Models concluding with convolutional layers demonstrated consistent performance across different quantisation levels, with accuracy variations limited to $\pm$1.2\%. In contrast, dense layer configurations showed higher sensitivity to quantisation, with accuracy variations of up to $\pm$3.5\%. 

\subsubsection{Student Model Architecture}
\begin{figure}
    \centering
    \includegraphics[width=0.95\linewidth]{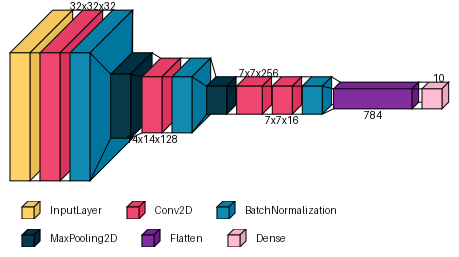}
    \caption{Student CNN model architecture using a traditional softmax classifier.}
    \label{fig:student_arch}
\end{figure}

The final student model is a significantly smaller convolutional neural network designed with ACAM deployment  as shown in Fig. \ref{fig:student_arch}. It consists of three main convolutional layers, where the first two are followed by batch normalisation and max pooling operations. The network begins with 32 filters in the first layer, increases to 128 in the second, and concludes with 256 filters in the third layer. An additional convolutional layer with 16 filters is appended to reduce the feature map size. The architecture has been designed to balance computational efficiency with model size and feature extraction capability.

To quantify the computational requirements of both architectures, we calculate the number of multiply-accumulate (MAC) operations. For each convolutional layer, the total number of MAC operations is given by:

\begin{equation}
\text{MACs}_{\text{conv}} = H_{\text{out}} \times W_{\text{out}} \times K_h \times K_w \times C_{\text{in}} \times C_{\text{out}}
\end{equation}
\noindent
where $H_{\text{out}}$ and $W_{\text{out}}$ are the output feature map dimensions, $K_h$ and $K_w$ are the kernel dimensions, and $C_{\text{in}}$ and $C_{\text{out}}$ are the input and output channels respectively. This calculation allows us to compare the computational complexity of our teacher and student models, demonstrating the efficiency gains achieved through our architectural choices.

\section{Results}
\subsection{Model compression performance}
The implementation of our compression methodology demonstrates a successful balance between model size reduction and maintained performance. As shown in Table \ref{tab:DNN_results}, our teacher model, based on the ResNet-50 architecture, achieved a baseline accuracy of 93.77\% on the CIFAR-10 dataset, with corresponding F1-score, precision, and recall metrics all around 93\%, which is similar to previous work \cite{Kaushik2024}. This performance comes at the cost of substantial computational requirements, with the model containing approximately 26.2 million parameters and requiring roughly 3 billion operations. The same teacher model trained on the greyscale CIFAR-10 dataset requires a similar number of parameters but achieves a slightly reduced accuracy of 91.04\% due to the reduced number of features. This demonstrates the increased challenge of working with greyscale images for the compressed student models.

\begin{table*}[t]
    \centering
    \begin{tabular}{ccccccccc}
        Model & Accuracy  & F1-Score  & Precision  & Recall & Parameters  & MAC operations & Compression ratio \\
        Teacher colour & 93.77 & 93.78 & 93.83 & 93.77 & 26,215,810 & 3,858,551,808 & -\\ 
        Teacher greyscale & 91.04 & 91.05  & 91.13  & 91.05 & 26,209,538 & 3,808,375,808 & 1.01:1 \\
        Student (without optimisations) & 76.29 & 76.29 & 76.14 & 79.29 & 380,314 & 23,785,120 & 162:1 \\
        Student (with optimisations) & 82.22 & 82.20  & 82.34  & 82.23 & 380,314  & 4,757,024 & 811:1 \\
        
    \end{tabular}
    \caption{Comparison of teacher and student model performance, calculated using a softmax layer for classification.}
    \label{tab:DNN_results}
\end{table*}


On the other hand, the unoptimised base student model, while achieving a significant reduction in model size to just 380,314 parameters (a 98.5\% reduction), initially showed a considerable performance drop to 76.29\% accuracy. This model required nearly 24 million operations, already representing a 99\% reduction in computational complexity compared to the teacher model. However, the application of our optimisation techniques, including knowledge distillation, pruning and quantisation significantly improved the student model's performance while also further reducing the number of operations. The optimised student model achieved an accuracy of 82.22\%, representing a substantial 5.93\% improvement over its unoptimised comparative model. This optimisation came with an additional benefit of reducing the effective MAC operations to 4.76 million through 80\% sparsity, as operations involving pruned (zero-valued) weights can be skipped entirely, representing an 800-fold reduction in operations compared to the original teacher model. By eliminating these unnecessary computations, the network achieves significant energy savings while maintaining classification accuracy.

While the performance gap between the teacher and optimised student model remains at 8.82\%, it should be considered in the context of the dramatic reduction in model complexity. The optimised student model maintains this performance level while using just 1.45\% of the parameters of the teacher model and requiring only 0.13\% of the computational operations. This trade-off between performance and resource efficiency positions the model favourably for deployment on resource-constrained ACAM hardware.

\subsection{Pattern matching performance analysis}


\begin{figure}[h]
    \centering
    \includegraphics[width=0.95\linewidth]{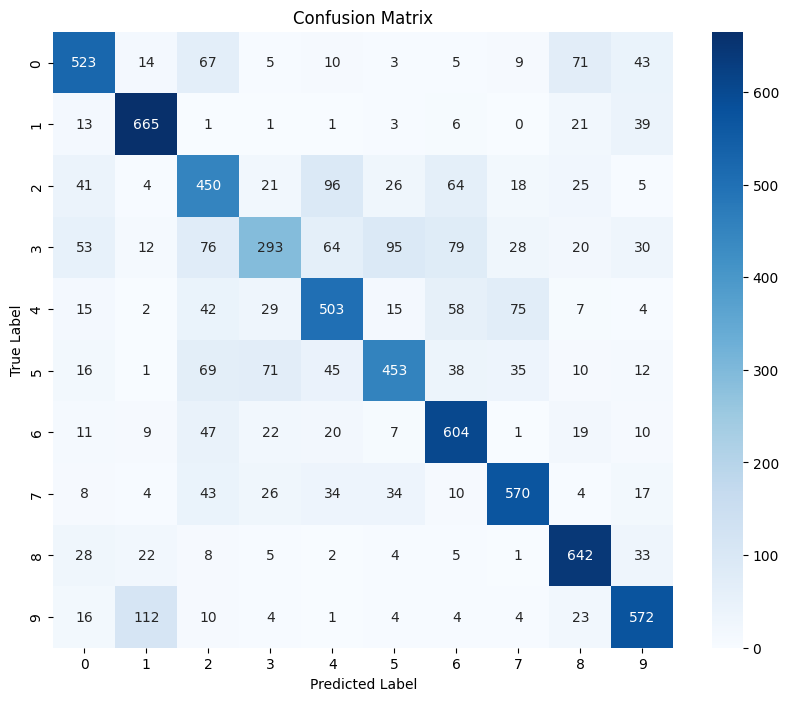}
    \caption{Confusion matrix of optimised student model using feature-based pattern matching classifier.}
    \label{fig:cf}
\end{figure}

The evaluation of different pattern matching approaches reveals important insights about the trade-offs between accuracy and computational complexity. Binary (1-bit) template quantisation achieved consistent performance of 70.91\% accuracy across both feature count (see Fig. \ref{fig:cf}) and similarity-based approaches, though this represents a reduction from the 82.22\% accuracy achieved by the student model using a traditional softmax classifier. This ~11\% accuracy drop must be considered in the context of the significant computational benefits - while the softmax classifier requires floating-point operations and dense matrix multiplications, the binary pattern matching approaches operate using only simple bit comparisons.

The identical performance between feature count and similarity-based approaches in the binary domain can be explained by examining how the matching operations converge when features are quantised to single bits. In the binary case, both methods effectively reduce to counting matching bit positions, since the similarity approach's distance calculations and hit ratio measurements become equivalent to direct bit comparisons when working with binary values. This convergent behaviour suggests that more complex similarity calculations provide no additional benefit when working with these features.

While binary pattern matching achieves lower accuracy compared to the student model with softmax, its computational simplicity and minimal memory requirements make it an attractive option for extremely resource-constrained edge applications where the ~11\% accuracy trade-off is acceptable in exchange for significant energy savings and hardware simplification.


\begin{figure}[h]
    \centering
    \includegraphics[width=0.95\linewidth]{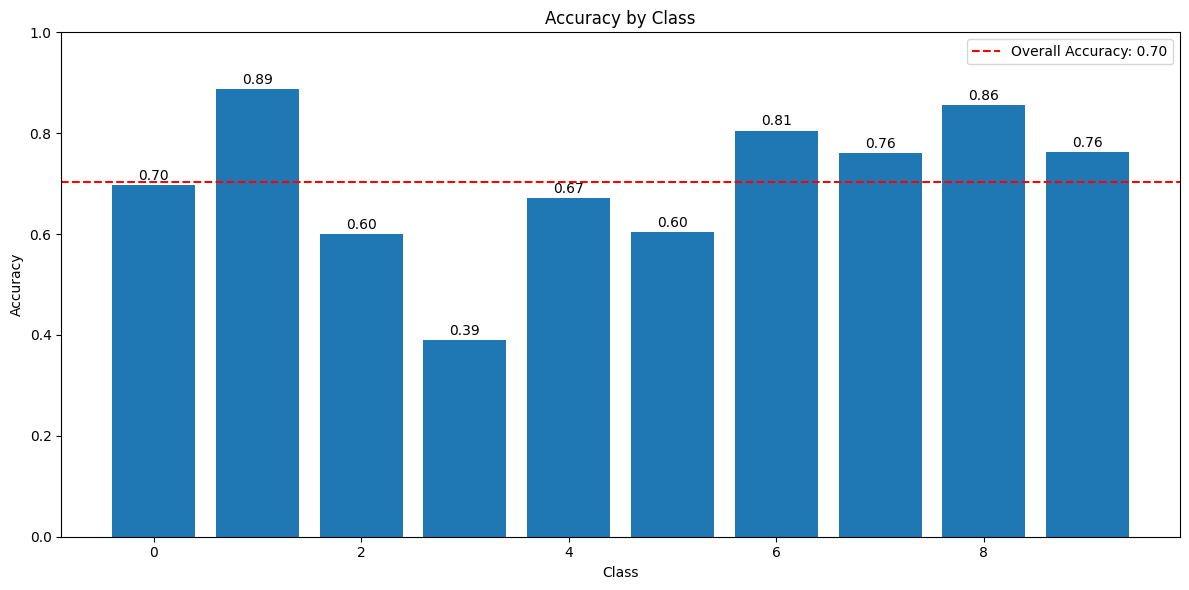}
    \caption{Per-class accuracy of optimised student model using feature-based pattern matching classifier.}
    \label{fig:acc}
\end{figure}

\subsection{Multiple template analysis}
The investigation of using multiple templates per class reveals slight improvements in classification accuracy, as shown in Table \ref{tab:templates}. This analysis explores configurations using one, two, and three templates per class to capture intra-class variations more effectively.


\begin{table}[h]
    \centering
    \begin{tabular}{cccc}
        Number of templates & Accuracy (\%) \\
         1 & 70.91   \\
         2 & 71.64  \\
         3 & 71.60   \\
    \end{tabular}
    \caption{Comparison of accuracy when using varying number of templates.}
    \label{tab:templates}
\end{table}

The increasing the number of templates from one to two improved accuracy from 70.91\% to 71.64\%, a gain of 0.73\%. However, further increasing to three templates per class showed a slight decrease to 71.60\%, suggesting diminishing returns from additional templates. This pattern suggests that binary quantisation, with its reduction of features, creates relatively distinct and well-separated class representations where a single template can already capture the essential pattern for each class. The minimal improvement with a second template and decrease with a third indicates that these features lack the granularity to meaningfully distinguish additional intra-class patterns, essentially leading to redundant or potentially conflicting templates.


These consistently small improvements indicate that increasing the number of templates may not be an effective strategy for enhancing classification performance. The limited gains must be weighed against the significant increase in memory requirements and computational complexity associated with storing and comparing multiple templates per class. This suggests that simpler configurations with fewer templates might offer a better balance between resource utilisation and classification performance. The results indicate that focusing on optimising the quality of a single template per class might be more beneficial than increasing the number of templates, particularly given the minimal returns observed with additional templates.




\subsection{Energy efficiency estimates}
For the back-end classifier, each template matching operation consumes 185fJ per cell. With our architecture requiring 10 templates of 784 features each, the energy consumption is:

\begin{equation}
\begin{split}
E_{\text{back-end}} &= N_{\text{templates}} \times N_{\text{features}} \times E_{\text{cell}} \\
&= 10 \times 784 \times 185\text{fJ} = 1.45\text{nJ}
\end{split}
\end{equation}

The total energy consumption of our front end network can be calculated considering both sparsity and quantisation effects. Given our total MAC operations of 23,785,120 from Table \ref{tab:DNN_results}, and accounting for 80\% sparsity, the effective number of MAC operations is 4,757,024. However, we must also consider that lack of the final softmax fully-connected layer using this ACAM optimised approach, therefore we can remove an addition 7,850 operations that were used by this final layer, meaning the total operations for the front end classifier is 4,749,174. Using the energy figures from Horowitz \cite{Horowitz2014} for 8-bit operations (0.2pJ for multiply and 0.03pJ for add) and memory access costs (20pJ for 32KB cache), we can calculate the energy consumption of the front-end feature extractor. For each MAC operation, the computation energy is 0.23pJ and the memory access energy is 20pJ, giving a total front-end energy consumption of 96.07nJ per inference. This combines with our back-end ACAM energy of 1.45nJ for template matching, resulting in a total system energy consumption of 97.52nJ per classification operation. In comparison the teacher model has a total energy consumption of 78.06$\mu$J demonstrating the proposed methodology achieved a 792 times energy reduction.

This demonstrates that our hybrid approach achieves significant energy efficiency, with the ACAM back-end consuming less energy than traditional digital implementations. This paves the way for new opportunities in multi-modal sensing enabling smaller, more efficient deep learning models that can be implemented in devices such as wearables while preserving battery life.

\section{Conclusion}
This work demonstrates the feasibility of a hybrid digital-analogue architecture for edge image classification through the integration of optimised neural networks with RRAM-based ACAM hardware accelerators. The proposed digital methodology utilising knowledge distillation achieves substantial efficiency improvements while maintaining accuracy, with our resultant student model achieving 82.22\% accuracy on CIFAR-10 using only 1.45\% of the teacher model's parameters and reducing MAC operations by a factor of 800. For analogue classification, template pattern matching maintains 70.91\% accuracy using a simple feature count approach, suggesting more complex pattern matching approaches may not yield proportional benefits in classification accuracy.

This approach demonstrates significant advantages in terms of energy efficiency, with the ACAM back-end consuming only 1.45nJ per classification operation compared with 78.06$\mu$J for the teacher model, while maintaining acceptable accuracy levels for edge applications. These results, combined with our investigation of multiple template strategies, establish the potential of digital-analogue approaches for resource-constrained edge devices. Future research could further explore template generation techniques and novel ACAM cell designs, further optimising classification accuracy and energy efficiency in edge computing applications.

\bibliographystyle{IEEEtran}
\bibliography{references.bib, References_v2025_01_28_e4aGP.bib, papandroulidakis2024_9T4R_arXiv.bib}

\end{document}